\documentclass[letterpaper, 10pt, conference]{ieeeconf}

\usepackage{graphicx,amsmath,amssymb,booktabs,multirow,xcolor,url}
\usepackage[colorlinks=true,allcolors=blue]{hyperref}
\usepackage{capt-of}

\title{VOIM: Training-Free Open-Vocabulary 3D Instance Mapping for RGB-D and Monocular SLAM}
\IEEEoverridecommandlockouts
\author{Sangmin Song$^{1}$, Sarath Kodagoda$^{1}$, Marc G. Carmichael$^{1}$, Karthick Thiyagarajan$^{2}$,\\
Amal Gunatilake$^{1}$, Kelly Prentice$^{3}$, Jodi Martin$^{3}$
\thanks{This research was supported by the Australian Government through the
Australian Research Council's Linkage Projects funding scheme (LP220100430) and
the industry partner Guide Dogs NSW/ACT.}
\thanks{$^{1}$S. Song, S. Kodagoda, M. G. Carmichael and A. Gunatilake are with the
Robotics Institute, Faculty of Engineering and Information Technology, University of
Technology Sydney, Ultimo NSW 2007, Australia.
{\tt\small Sangmim.Song@student.uts.edu.au}}
\thanks{$^{2}$K. Thiyagarajan is with the Smart Sensing and Robotics Laboratory
(SensR Lab), Centre for Advanced Manufacturing Technology, School of Engineering,
Design and Built Environment, Kingswood NSW 2747, Australia.}
\thanks{$^{3}$K. Prentice and J. Martin are with Guide Dogs New South Wales/Australian
Capital Territory, Sydney NSW, Australia.}}

\makeatletter
\let\@oldmaketitle\@maketitle
\renewcommand{\@maketitle}{\@oldmaketitle\vspace{-1mm}%
  \begin{center}\includegraphics[width=\textwidth]{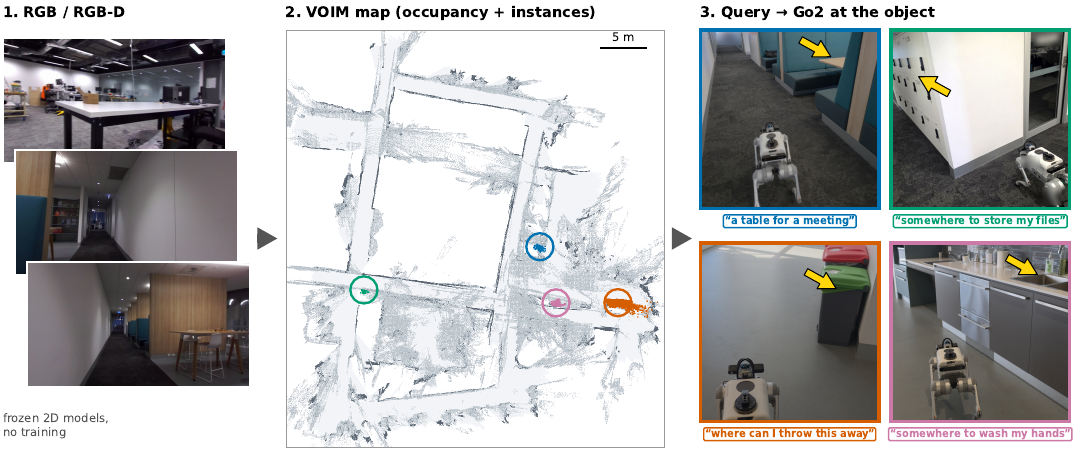}\end{center}\vspace{-4mm}%
  {\footnotesize\setcounter{figure}{0}\captionof{figure}{\textbf{From one mapping sweep to a robot at the requested object.} (1)~One hand-held RGB or RGB-D sweep,
mapped with frozen 2D models. (2)~The VOIM map of a building floor: occupancy grid and open-vocabulary instances.
(3)~Each everyday request selects one instance by top-1 retrieval and a Unitree Go2 navigates
there.}\label{fig:hero}}\vspace{1mm}}
\def\bstctlcite{\@ifnextchar[{\@bstctlcite}{\@bstctlcite[@auxout]}}
\def\@bstctlcite[#1]#2{\@bsphack\@for\@citeb:=#2\do{\edef\@citeb{\expandafter\@firstofone\@citeb}%
  \if@filesw\immediate\write\csname #1\endcsname{\string\citation{\@citeb}}\fi}\@esphack}
\makeatother
\begin{document}
\bstctlcite{IEEEexample:BSTcontrol}
\maketitle
\setcounter{figure}{1}

\begin{abstract}
Robots that act on everyday language requests need maps that record what each object is and where it is. Existing open-vocabulary 3D mapping systems usually require a depth sensor and commit to each object the first time they see it, when the evidence is weakest. We present VOIM, a training-free system that builds open-vocabulary 3D instance maps from RGB-D or from monocular RGB video alone. VOIM accumulates soft label evidence from frozen, off-the-shelf perception models in every voxel across views, and decides labels and object instances only after that evidence has built up. On ScanNet++, VOIM achieves the highest present-class mIoU among the compared methods, 10 points above the strongest baseline. From monocular video alone, VOIM also exceeds HOV-SG and ConceptGraphs run with depth and ground-truth poses. We also conduct real-robot navigation experiments on a quadruped to show that VOIM maps are usable in the real world.
\end{abstract}

\begin{keywords}
Semantic Scene Understanding, Mapping, RGB-D Perception, Vision-Based Navigation
\end{keywords}

\section{Introduction}\label{sec:intro}
\begin{figure*}[t]
  \centering
  \includegraphics[width=\textwidth]{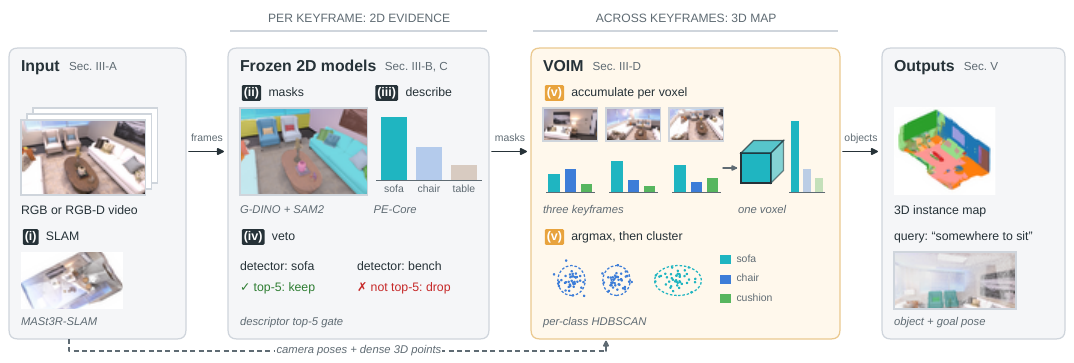}
  \caption{\textbf{System overview} on Replica \texttt{room0}. (i)~SLAM recovers camera poses and
  dense 3D points from RGB or RGB-D video. Per keyframe, (ii)~Grounding DINO and SAM2 propose labeled
  masks, (iii)~PE-Core scores each mask over the full vocabulary, and (iv)~the detector label is kept
  only if it is in the descriptor's top~5. (v)~VOIM sums these soft votes in every voxel across
  keyframes, takes the argmax only afterwards, and clusters each class into 3D instances. The map
  exports a per-instance text index and an occupancy grid (Sec.~\ref{sec:robot}). Bar heights are
  schematic.}
  \vspace{-2mm}
  \label{fig:pipeline}
\end{figure*}
A robot that is asked to ``find somewhere to wash my hands'' needs a map that links language
to objects in 3D. Open-vocabulary mapping methods build such maps by lifting vision--language
features from images into 3D~\cite{peng2023openscene,jatavallabhula2023conceptfusion,kerr2023lerf},
and recent systems do so online~\cite{martins2025ovoslam,werby2024hovsg}. Almost all of them
assume RGB-D input and known camera poses. Depth sensors add cost, fail outdoors and on reflective
surfaces, and are missing from most cameras that robots and phones already carry.

A second problem is when the map decides what an object is. Single views are unreliable:
detectors hallucinate classes, and masks bleed across depth discontinuities, which is also where
reconstruction is weakest. Recent online systems such as OVO-SLAM~\cite{martins2025ovoslam} brought
open-vocabulary mapping into full SLAM pipelines, but they create each object instance at its first
detection and only refine it afterwards, so early segmentation errors carry forward.
Closed-set semantic mapping addressed this long ago by fusing labels over many
views~\cite{mccormac2017semanticfusion}; we bring that idea to open vocabularies.

VOIM stores soft class distributions in every voxel and updates them from each new view, using
frozen detection, segmentation, and region-description
models~\cite{liu2023groundingdino,ravi2024sam2,xiao2025textregion,bolya2025pe} as released. A
cross-model veto suppresses classes that the detector and the descriptor disagree on. Labels are
read out only after accumulation, and object instances are formed afterwards by clustering points
of each class in 3D. The same pipeline accepts posed RGB-D or plain RGB video. For RGB, a
feed-forward SLAM backbone~\cite{murai2025mast3rslam,maggio2025vggtslam} supplies poses and
depth. Other training-free monocular object maps~\cite{kassab2026lexisg,yu2026ranger} do not report
the 3D semantic accuracy of the map. SLAM runs at its own rate while semantics are added
behind it; the semantic stream is the slow part, mainly because the detector runs once per class.

We compare against the baselines on the same frames, poses, and full class vocabulary, and report
both per-scene and pooled mIoU (Sec.~\ref{sec:protocol}). On ScanNet++~\cite{yeshwanth2023scannetpp}
with RGB-D and ground-truth poses, VOIM achieves the highest present-class mIoU, computed over the classes
that occur in each scene, among the compared methods, 42.3 per scene against 32.4 for the strongest baseline, OVO-SLAM, and is ahead in all ten
scenes.
Removing the detector costs VOIM about five points but keeps it ahead in nine of ten scenes,
which isolates the contribution of the mapping stage. On Replica~\cite{straub2019replica}, VOIM
with RGB-D is ahead on pooled mIoU and on par per scene. From RGB
video alone, VOIM reaches 24.78 per-scene mIoU on ScanNet++, above HOV-SG and ConceptGraphs,
which both use depth and ground-truth poses. Under all-classes scoring, which also penalizes
predicted classes that are absent from a scene, OVO-SLAM stays ahead, a trade-off we analyze in
Sec.~\ref{sec:scannetpp}.

Our contributions are:
\textbf{(i)}~VOIM, a training-free open-vocabulary instance mapper that accumulates soft per-voxel
evidence across views, fuses detector and descriptor outputs with a cross-model veto, and defers
both label and instance decisions until after accumulation;
\textbf{(ii)}~a controlled comparison with matched inputs and full-vocabulary prompting, in which
VOIM wins every ScanNet++ scene and a detector-free variant shows that the mapping
stage itself accounts for much of the gain;
\textbf{(iii)}~to our knowledge, the first training-free system to report instance-level 3D
semantic accuracy from monocular video, whose ScanNet++ maps exceed HOV-SG and ConceptGraphs run
with depth and ground-truth poses; and
\textbf{(iv)}~maps that support language-goal navigation, with an occupancy grid and a per-instance
text index, evaluated on 1,104 retrieval queries and demonstrated on a Unitree Go2 quadruped that
reached a fitting object for nine of twelve everyday requests across a building floor
(Fig.~\ref{fig:hero}).

\begin{figure*}[t]
  \centering
  \includegraphics[width=\textwidth]{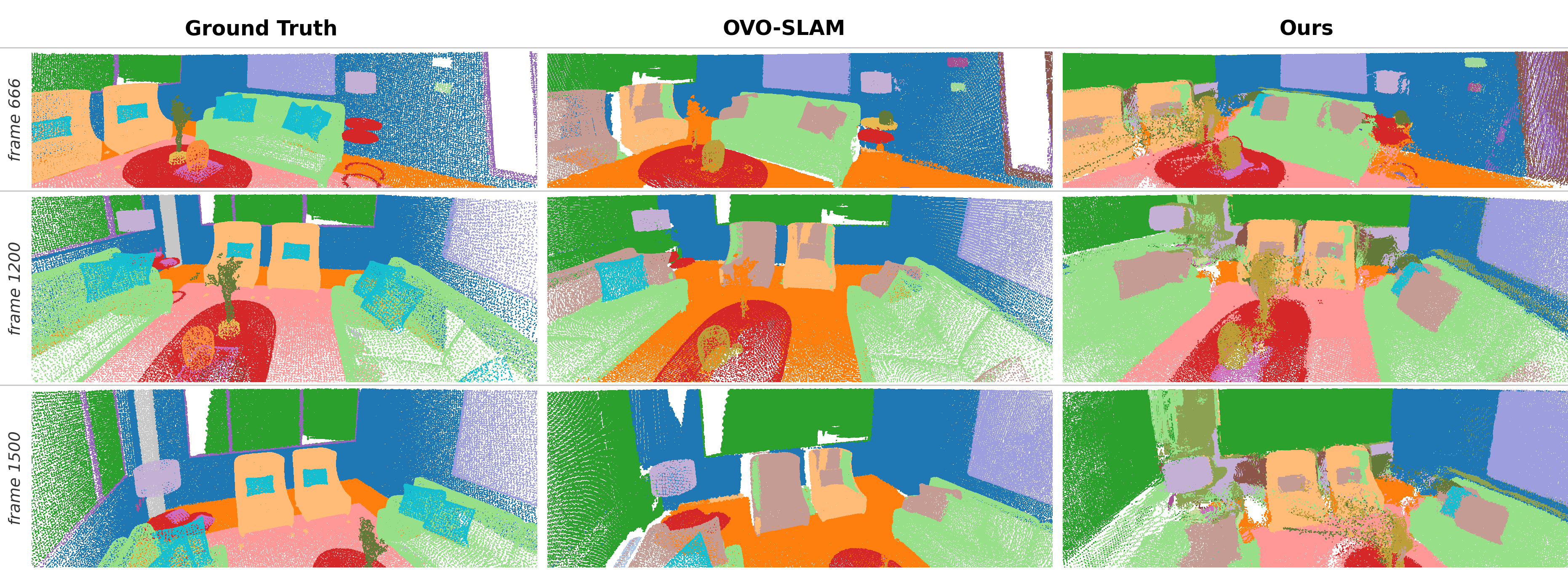}\\[2pt]
  \includegraphics[width=0.85\textwidth]{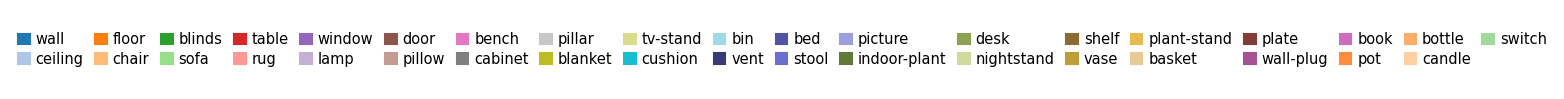}
  \caption{\textbf{Qualitative comparison on Replica \texttt{room0}.} Columns show the ground-truth
  semantics, OVO-SLAM from RGB-D and ground-truth poses, and VOIM from monocular video alone, without
  depth or poses. Each row renders the three point clouds from the same camera pose with a shared
  palette. The monocular map recovers the furniture and main surfaces, and most visible errors are
  wrong labels rather than wrong geometry.}
  \label{fig:qualitative}
\end{figure*}

\section{Related Work}\label{sec:related}
\textbf{Semantic mapping by multi-view label fusion.}
Multi-view label accumulation~\cite{mccormac2017semanticfusion,narita2019panopticfusion,rosinol2020kimera},
volumetric instance-aware mapping and 3D discovery~\cite{grinvald2019volumetric}, and Fusion++
object SLAM~\cite{mccormac2018fusionpp} predate open vocabularies. VOIM builds on this idea; what is new is what each voxel stores and
when decisions are made. Each voxel accumulates detector-gated vision--language distributions over
the full open vocabulary, and both the label and the object instance are decided only after
accumulation. Sec.~\ref{sec:scannetpp} shows that this mapping stage, not any single perception
model, carries much of the gain.

\textbf{Open-vocabulary 3D scene understanding.}
Offline methods work on posed RGB-D or a reconstructed scene: they attach pixel-aligned
CLIP~\cite{radford2021clip} features to 3D points~\cite{peng2023openscene}, fuse multimodal features
into an RGB-D map~\cite{jatavallabhula2023conceptfusion}, or embed language fields in a
NeRF~\cite{kerr2023lerf}. OpenMask3D, OVIR-3D, and Open3DIS first segment class-agnostic 3D proposals
and then label them~\cite{takmaz2023openmask3d,lu2023ovir3d,nguyen2024open3dis}, an offline
segment-then-label order that VOIM reverses. Open-vocabulary detectors, promptable
segmenters~\cite{liu2023groundingdino,ravi2024sam2}, and region
descriptors~\cite{bolya2025pe,xiao2025textregion} supply the 2D evidence. Combining a detector label
with an independent classifier score follows score-level fusion in open-vocabulary
detection~\cite{kuo2023fvlm,kaul2023multimodal}, which we adapt to per-voxel accumulation in 3D.

\textbf{Online open-vocabulary mapping and SLAM.}
ConceptGraphs builds object graphs from SAM masks~\cite{gu2024conceptgraphs}, Open-Fusion a
real-time queryable TSDF~\cite{yamazaki2024openfusion}, and HOV-SG hierarchical scene graphs from
RGB-D~\cite{werby2024hovsg}; our instance export follows this line of language-queryable maps for
navigation~\cite{huang2023vlmaps,shafiullah2023clipfields}. OVO-SLAM~\cite{martins2025ovoslam} tracks
class-agnostic 3D segments online, describes each with CLIP features merged across views, and runs
inside full SLAM pipelines with loop closure. Its segments are created at first detection and
tracked afterwards, whereas VOIM forms instances only after all evidence is in.

\textbf{Concurrent work.}
OVI-MAP~\cite{deng2026ovimap} and OpenVox~\cite{openvox2025} are training-free online instance
mappers but require RGB-D and supplied poses. Ov3R~\cite{gong2026ov3r} is monocular but trained,
reconstructs rather than performs SLAM, and produces no instances. KM-ViPE~\cite{kmvipe2025},
RADIO-ViPE~\cite{radiovipe2026}, and OpenMonoGS-SLAM~\cite{yoo2025openmonogs} produce no instances
either: the first two build semantic maps only, and OpenMonoGS-SLAM pairs MASt3R with Gaussian-splat
semantics and reports prompt-IoU. Training-free monocular object mapping has also appeared:
LEXI-SG~\cite{kassab2026lexisg} builds room-wise scene graphs from feed-forward reconstruction but is
scored with GT poses and depth, and RANGER~\cite{yu2026ranger} pairs MASt3R-SLAM with Grounding DINO
and SAM for object navigation without reporting map accuracy. MonoEM-GS~\cite{kruzhkov2026monoemgs}
reports monocular Replica mIoU per Gaussian rather than per instance, IRIS-SLAM~\cite{xiao2026irisslam}
trains its instance head, and CAMFusion~\cite{martins2026camfusion} learns multi-view descriptor fusion
for OVO-SLAM; their Replica numbers appear in Table~\ref{tab:replica} for context. Closest to our
deferral, FUS3DMaps~\cite{homberger2026fus3dmaps} fuses voxel and instance layers only at query time,
and OpenBelief-Nav~\cite{nguyen2026openbeliefnav} keeps per-object label beliefs and finds early
commitment worse. Merge vetoes and critics~\cite{hao2026ovipsg,bickici2026thinkwhileyoumap} gate
instance merges, whereas our veto gates labels across perception models. DualMap~\cite{jiang2025dualmap}
navigates a quadruped to language goals from an RGB-D map.

\textbf{Feed-forward monocular SLAM.}
Classical monocular SLAM is sparse and scale-ambiguous, hindering dense semantic fusion.
Feed-forward DUSt3R regresses dense two-view pointmaps~\cite{wang2024dust3r}, MASt3R adds
matching and metric-grade geometry~\cite{leroy2024mast3r}, and MASt3R-SLAM provides our primary
real-time dense monocular front-end~\cite{murai2025mast3rslam}. VGGT processes frame sets in
one pass~\cite{wang2025vggt}, with VGGT-SLAM extending it to submap-based dense
SLAM~\cite{maggio2025vggtslam}.

\section{Method}\label{sec:method}
\subsection{Overview}\label{sec:overview}
VOIM turns posed RGB-D frames, or monocular RGB video, into a dense point map in which every point
carries a distribution over open-vocabulary classes and belongs to a 3D object instance. With
monocular input, the map is defined up to a similarity transform. Fig.~\ref{fig:pipeline} shows
the five stages on monocular Replica \texttt{room0}. (i)~A SLAM backend,
MASt3R-SLAM~\cite{murai2025mast3rslam} or, where noted, VGGT-SLAM~\cite{maggio2025vggtslam},
provides camera poses and a dense pointmap for each keyframe. (ii)~Grounding
DINO~\cite{liu2023groundingdino} and SAM2~\cite{ravi2024sam2} propose labeled object masks.
(iii)~PE-Core with TextRegion~\cite{xiao2025textregion,bolya2025pe} scores each mask over the full
vocabulary. (iv)~A cross-model veto fuses the detector and descriptor outputs
(Sec.~\ref{sec:fusion}). (v)~VOIM accumulates these distributions in the map and extracts
instances (Sec.~\ref{sec:voim}).

\subsection{Perception}\label{sec:perception}
\textbf{Detection.} The semantic stream runs alongside tracking and labels keyframes as fast as it
can, which on Replica is 14--22 of about 66 keyframes per scene (Sec.~\ref{sec:runtime}). For each
keyframe, we prompt Grounding DINO (Swin-B) once per class with ``a [label]'' and keep boxes above
confidence 0.25. Prompting one class at a time avoids the truncation and order bias of a single long
caption listing the whole vocabulary, and needs no knowledge of which classes are in the scene.
Detector inputs are at full resolution on Replica and 512 pixels on ScanNet++. SAM2 (Hiera-L) turns
each box into a mask; we request several candidates and keep the highest-scoring one. Detections
from all class prompts are kept together, and when two masks with the same label overlap by more
than 0.7 IoU, the less confident one is dropped. Each mask $m$ therefore has one detector label
$\ell^{\mathrm{det}}_{m}$ and confidence $c_m$, while masks with different labels may overlap.

\textbf{Region descriptor.} Separately, TextRegion~\cite{xiao2025textregion} pools frozen PE-Core
L14-336 features inside each mask, giving one embedding per mask in the shared image--text space.
A softmax over its cosine similarities to the class text embeddings, with temperature
$\tau{=}0.03$, gives a distribution $p_m$ over the vocabulary $\mathcal{Y}$. Each class text
embedding averages the 80 standard CLIP ImageNet prompt templates~\cite{radford2021clip}, the
default in PE-Core's released classification recipe; we did not tune the prompts. Neither model is
fine-tuned or re-weighted.

\subsection{Label fusion with a cross-model veto}\label{sec:fusion}
The two sources fail in different ways. The detector is precise on objects that are present but
also fires on classes that are absent; the descriptor is less confident but gives a distribution for
every mask. We combine them per mask:
\begin{equation}
  \tilde{p}_{m} \;=\; \alpha\,\mathbf{1}[\ell^{\mathrm{det}}_{m}] \;+\; (1-\alpha)\,p_{m},
  \qquad \alpha=0.5,
  \label{eq:fusion}
\end{equation}
where $\mathbf{1}[\ell^{\mathrm{det}}_{m}]$ is a vector over the vocabulary with 1 at the detector's
class and 0 elsewhere, so $\tilde p_m$ gives half its weight to the detector's label and half to the
descriptor's distribution. The detector label therefore always wins the per-mask $\arg\max$. To stop it from overriding the
descriptor on absent classes, we accept it only if the descriptor also ranks that class in its top 5; otherwise the mask keeps the descriptor distribution $p_m$ alone. This fixed
gate uses no ground truth or scene information, in line with how frozen image--text scores are
tempered in open-vocabulary detection~\cite{kuo2023fvlm,kaul2023multimodal}. The veto is used in all
main results. The ScanNet++ perception controls (Sec.~\ref{sec:scannetpp}) start from a
configuration without it, and Sec.~\ref{sec:ablation} ablates it.

\subsection{VOIM: Voxel-Grounded Incremental Instance Manager}\label{sec:voim}
VOIM lifts the per-mask evidence into a 3D map organized by instances. Like SemanticFusion-style
mapping, it keeps soft label distributions and takes the argmax only after all views are in. Unlike
online open-vocabulary baselines, which associate 2D masks with 3D instances as they arrive and later
merge and relabel them, it also defers the instance decision until after accumulation.

\textbf{Voxel map.} Keyframe pointmaps are transformed into a common frame with their poses and
merged on a regular voxel grid. Each occupied voxel stores the average position and color of its
points, and remembers every (keyframe, pixel) pair that fell into it, and hence every mask that
observed it.

\textbf{Soft label voting.} Each point $v$ sums the fused distributions of all masks that observed
it, weighted by detector confidence:
\begin{equation}
  P_{v}(\ell) \;=\; \frac{1}{Z_{v}} \sum_{m\,\in\,\mathcal{O}(v)} c_{m}\;\tilde{p}_{m}(\ell),
  \qquad
  Z_{v} = \sum_{m\,\in\,\mathcal{O}(v)} c_{m},
  \label{eq:voting}
\end{equation}
where $\mathcal{O}(v)$ is the set of masks that observed $v$. The total weight $Z_v$ also serves as
a per-point confidence. Because this is a running sum, a new keyframe only adds votes and never
revisits earlier decisions. Alongside the labels, each point averages the normalized region
embeddings of its masks; text queries are matched against this per-point feature
(Sec.~\ref{sec:robot}). Sec.~\ref{sec:scannetpp} and Sec.~\ref{sec:ablation} measure the effect of
the mapping stage and of soft voting, and Table~\ref{tab:instances} measures the deferred instance
decision.

\textbf{Instance extraction.} After accumulation, each point takes its most likely label,
$\hat\ell_v=\arg\max_\ell P_v(\ell)$, and a majority vote over its 15 nearest neighbors, requiring
60\% agreement, smooths labels between neighboring points. For each class, HDBSCAN then clusters that
class's points into instances. This separates repeated objects, such as two chairs, and marks scattered false
positives as noise. Each instance stores its class, its averaged open-vocabulary feature, and its
points. On ScanNet++ we score the per-point labels before clustering, and on Replica the clustered
instances. VOIM runs as two streams. The SLAM backend tracks the camera and builds the
dense geometry at its own rate. A separate semantic stream detects, describes, and votes on
keyframes as fast as it can, skipping those it cannot keep up with. Because labels and instances
are decided only after accumulation, semantics that arrive late cost nothing: they simply add votes
to points that already exist. The final argmax and clustering are a short batch step at the end. In
this paper, ``online'' therefore means that the map is built incrementally in a single pass; the
semantic stream is the slow part and runs well below real time (Sec.~\ref{sec:runtime}).
\section{Experiments}\label{sec:exp}

\subsection{Evaluation protocol}\label{sec:protocol}
We evaluate on ScanNet++ (10 scenes, top-100 classes) and Replica (8 scenes, 51 classes).
Predictions are transferred to the ground-truth mesh by giving each mesh vertex the majority label
of its five nearest predicted points, with no distance cutoff, and scored as mIoU over the benchmark
vocabulary. Classes absent from both prediction and ground truth are skipped, and a class that is
predicted but absent from the scene scores zero. On ScanNet++, the present classes are those that
occur in the scene's ground truth. On Replica we use OVO-SLAM's evaluation code~\cite{martins2025ovoslam},
and on ScanNet++ our own implementation of the same procedure. We call VOIM run on the same RGB-D
frames and ground-truth poses as the baselines \emph{input parity}, and VOIM run on RGB video alone
\emph{monocular}.

We take four steps to keep the comparison fair. First, every method receives the full vocabulary,
never a scene-specific class list. Second, our evaluation reproduces OVO-SLAM's published Replica
result with its released models (27.53 vs.\ 27.0 pooled). Third, frame schedules are matched: on
ScanNet++ all methods use the same frames, and OVO-SLAM maps and segments every one of them, the best
of seven schedules we tried; on Replica it segments about 200 of 2000 frames, input-parity VOIM labels
100, and monocular VOIM 14--22 SLAM keyframes. Fourth, we report two aggregations: pooled mIoU, which
sums confusion matrices over scenes, and the per-scene mean. One asymmetry remains: on ScanNet++,
OVO-SLAM receives depth rendered from the mesh while VOIM uses the sensor depth, which favors the
baseline. On Replica, ``RGB-D'' always means depth rendered from the mesh.

\emph{Monocular alignment.} A monocular map is defined only up to scale, rotation, and translation,
so we align it to the ground truth with a single similarity transform, fitted between the estimated
and ground-truth camera trajectories~\cite{umeyama1991} and applied unchanged to the map. The fit uses
geometry only, never labels, and has seven parameters, whereas the baselines receive ground-truth
poses for every frame. Input-parity maps use ground-truth poses and need no alignment.

\emph{Settings.} All runs use the frozen models and settings of Sec.~\ref{sec:method}. Only three
settings differ between datasets: voxel size (2\,cm on ScanNet++, 1\,cm on Replica), detector input
(512 pixels on ScanNet++, full resolution on Replica), and what is scored (per-point labels on
ScanNet++, clustered instances on Replica). The veto is on for both datasets in our main
configuration; a per-dataset variant turns it off on ScanNet++, which raises present-class mIoU there
but predicts more absent classes (Sec.~\ref{sec:scannetpp}).

\begin{table}[t]
\centering
\caption{Replica semantic segmentation (8 scenes, 51 classes, mIoU).}
\label{tab:replica}
\setlength{\tabcolsep}{2.5pt}\footnotesize
\begin{tabular}{lcccc}
\toprule
Method & Depth & Pose & pooled & per-scene \\
\midrule
HOV-SG~\cite{werby2024hovsg}$^\S$ & RGB-D & GT & --- & 22.5 \\
Open3DIS~\cite{nguyen2024open3dis}$^\S$ & RGB-D & GT & --- & 25.6 \\
OVO-SLAM~\cite{martins2025ovoslam} (published)$^\S$ & RGB-D & GT & 27.0 & --- \\
OVI-MAP~\cite{deng2026ovimap}$^\S$ & RGB-D & GT & \multicolumn{2}{c}{26.5} \\
OVO (our re-run) & RGB-D & GT & 27.53 & 30.10 \\
ConceptGraphs~\cite{gu2024conceptgraphs} (our re-run) & RGB-D & GT & 7.24 & 10.39 \\
OVO w/ PE+TR~\cite{martins2026camfusion}$^\S$ & RGB-D & GT & \multicolumn{2}{c}{33.0} \\
CAMFusion~\cite{martins2026camfusion} (trained)$^\S$ & RGB-D & GT & \multicolumn{2}{c}{38.3} \\
MonoEM-GS~\cite{kruzhkov2026monoemgs}$^\S$ & RGB only & est. & \multicolumn{2}{c}{31.5} \\
IRIS-SLAM~\cite{xiao2026irisslam} (trained)$^\S$ & RGB only & est. & --- & 20.22 \\
\midrule
\textbf{Ours} (input parity) & RGB-D (GT) & GT & \textbf{35.62} & \textbf{30.81} \\
\textbf{Ours} (monocular) & RGB only & est. & 27.75 & 23.55 \\
\quad\emph{w/o veto (ablation)} & RGB only & est. & 25.78 & 19.48 \\
\bottomrule
\end{tabular}
\\[2pt]
\footnotesize $^\S$As reported, under protocols that differ from ours; for context only. Other
rows use OVO-SLAM's evaluation code and the full vocabulary.
\end{table}

\subsection{ScanNet++}\label{sec:scannetpp}
\begin{table}[t]
\centering
\caption{ScanNet++ semantic segmentation (10 scenes, top-100 classes, mIoU). RGB-D and GT
poses unless marked.}
\label{tab:scannetpp}
\setlength{\tabcolsep}{4pt}\footnotesize
\begin{tabular}{lccc}
\toprule
 & \multicolumn{2}{c}{present-classes} & all-classes \\
\cmidrule(lr){2-3}\cmidrule(lr){4-4}
Method & pooled & per-scene & per-scene \\
\midrule
HOV-SG~\cite{werby2024hovsg}$^\P$ & 14.26 & 17.25 & 12.49 \\
ConceptGraphs~\cite{gu2024conceptgraphs}$^\P$ & 8.07 & 8.27 & 5.10 \\
OVO-SLAM~\cite{martins2025ovoslam} (frame-matched)$^\dagger$ & 25.97 & 32.37 & \textbf{24.86} \\
\textbf{Ours} & \textbf{33.98} & \textbf{42.32} & 16.95 \\
\midrule
\quad\emph{per-dataset (veto off)} & 33.31 & 44.07 & 13.27 \\
\quad\emph{balanced}$^\|$ & 31.12 & 37.39 & 21.83 \\
\quad\emph{SigLIP descriptor}$^\S$ & 29.53 & 39.96 & --- \\
\quad\emph{descriptor labels, DINO masks} & 30.66 & 37.13 & --- \\
\quad\emph{descriptor labels, agnostic masks}$^\ddagger$ & 30.24 & 38.82 & --- \\
\midrule
Ours (monocular, RGB only)$^{*}$ & 17.87 & 24.78 & 10.82 \\
\bottomrule
\end{tabular}
\\[2pt]
\footnotesize $^\P$Our re-run with the authors' code on the same inputs. $^\dagger$Best of seven
schedules; 18.58 with its default. $^\ddagger$No detector: class-agnostic SAM2 masks labeled by the
descriptor. $^\S$SigLIP2 instead of PE-Core. $^\|$Stricter veto, fewer absent classes.
$^{*}$RGB only, MASt3R-SLAM poses.
\end{table}
Table~\ref{tab:scannetpp} gives the main result and the perception controls. HOV-SG and
ConceptGraphs, re-run on identical inputs, trail both online systems in every scene. ConceptGraphs
uses its class-agnostic setting (8.04 per scene with its default), HOV-SG needed two patches that
only fix crashes, and both are scored with background classes, which their own evaluations exclude.
VOIM is ahead of OVO-SLAM in every scene (sign test $p\approx0.002$), by 0.19 to 17.29 points. This
advantage is specific to present-class scoring; under all-classes scoring OVO-SLAM leads, as analyzed
below.

The perception controls ask whether the gain comes from better 2D models rather than from mapping.
They start from the per-dataset variant, which reaches 44.07 per scene. Removing the detector, so that
class-agnostic SAM2 masks are labeled by the descriptor alone, lowers it to 38.82, and replacing
PE-Core with a SigLIP2 descriptor from the baseline's model family lowers it to 39.96. Every control
stays 4.76--11.70 points ahead of OVO-SLAM, and the SigLIP2 variant still wins 8 of 10 scenes, losing
where the margin was thinnest. Of the 11.70-point margin, about 6.94 points come from the detector's
label prior, if the label and mask effects add.

Under all-classes scoring, which charges every predicted class that is absent from the scene, the
ranking flips: $16.95$ for VOIM against $24.86$, in all ten scenes. The cause is how many classes VOIM
predicts, not how well it segments. The per-dataset variant predicts $53.0$ absent classes per scene
against the baseline's $7.2$, because it takes a per-point argmax over the full vocabulary after
about 2,400 per-mask decisions, while the baseline labels 144 tracked instances. A tighter veto with
minimum class support, the balanced row, recovers part of the gap, but once absent-class counts match
the baseline, the present-class advantage is gone. The same effect explains the Replica deficit
(Sec.~\ref{sec:replica}).
No prior work reports this benchmark in 3D: OVO-SLAM evaluates only 2D descriptor merging on
ScanNet++, and concurrent systems report other metrics~\cite{gong2026ov3r}, so we re-run every
baseline whose code accepts posed RGB-D. The monocular row is discussed in Sec.~\ref{sec:decomp}.

\subsection{Replica: input parity and monocular}\label{sec:replica}

In Table~\ref{tab:replica}, input-parity VOIM is above the published OVO-SLAM result on pooled
mIoU and on par per scene (30.81 vs.\ 30.10). VOIM labels 100 frames per scene and the baseline
about 200; labeling 200 moves VOIM to 35.94 pooled and 31.25 per scene, so view count does not
explain the comparison. Monocular VOIM is level pooled (27.75 vs.\ 27.53) and trails per scene
(23.55 vs.\ 30.10). Its per-scene deficit is concentrated in small offices, worst on
office1 at $15.4$, where a few large-surface classes make each misassignment costly.
Open3DIS (25.6) labels an offline RGB-D reconstruction, and only input-parity VOIM exceeds it.
ConceptGraphs, re-run with the same class-agnostic setting as on ScanNet++, reaches 10.39 per scene.
Our HOV-SG re-run exhausted the 62\,GB of system memory on all eight scenes, so we list its reported
number.
Replica's all-classes score charges absent-class predictions and weights small offices as heavily
as large rooms, which exposes the detector failure described above; the mapping-stage result
therefore holds for ScanNet++ present-class scoring. Fig.~\ref{fig:qualitative} shows the monocular
reconstruction. Among the concurrent systems in Sec.~\ref{sec:related} and Table~\ref{tab:replica},
only VOIM reports monocular instance-level 3D mIoU without training.

\subsection{Instance-level evaluation}\label{sec:instances}
Because VOIM forms instances only after accumulation, we also evaluate the instances themselves
(Table~\ref{tab:instances}): greedy average precision at IoU 0.25 and 0.50 over the 51 classes,
coverage, the share of ground-truth instances matched at IoU${\ge}0.25$, and purity. Predicted points
are transferred to the ground-truth mesh by the same five-neighbor vote, here with a 10\,cm match
radius, and matched to Replica's object annotations; OVO-SLAM's instances are scored the same way.
\begin{table}[t]
\centering
\caption{Replica instance-level evaluation (8 scenes, 51 classes).}
\label{tab:instances}
\setlength{\tabcolsep}{3pt}\footnotesize
\resizebox{\columnwidth}{!}{%
\begin{tabular}{lcccccc}
\toprule
Method & Depth & Pose & AP@25 & AP@50 & Cov. & Purity \\
\midrule
OVO-SLAM & RGB-D & GT & 32.87 & 25.93 & 41.24 & 65.12 \\
\textbf{Ours} (input parity) & RGB-D (GT) & GT & 36.66 & 25.79 & 42.08 & 61.21 \\
\textbf{Ours} (monocular) & RGB only & est. & 27.28 & 14.79 & 30.1 & 52.4 \\
\bottomrule
\end{tabular}
}
\end{table}
At input parity, deferring the instance decision makes instances more complete: VOIM is ahead on
AP@25 (36.66 vs.\ 32.87) and coverage (42.08 vs.\ 41.24), level on AP@50 (25.79 vs.\ 25.93), and
slightly behind on purity (61.21 vs.\ 65.12). From monocular video, VOIM trails the RGB-D baseline by
5.6 AP@25 and 11.1 AP@50, with lower purity (52.4) and coverage (30.1\%), so about a quarter of the
instances OVO-SLAM recovers are missed entirely. Both methods split large surfaces such as ceilings
into dozens of instances, which lowers AP.

\subsection{Ablations}\label{sec:ablation}
\begin{table}[t]
\centering
\caption{Ablations on monocular Replica (8 scenes, 51 classes). Defaults in bold.}
\label{tab:ablations}
\setlength{\tabcolsep}{3pt}\footnotesize
\resizebox{\columnwidth}{!}{%
\begin{tabular}{lccc}
\toprule
Sweep & value & per-scene mIoU & pooled mIoU \\
\midrule
DINO-prior weight $\alpha$ & 0.3 / \textbf{0.5} / 0.7 & 23.87 / 23.55 / 23.56 & 27.73 / 27.75 / 28.07 \\
Veto depth $k$ & 3 / \textbf{5} / 10 & 23.33 / 23.55 / 22.93 & 27.29 / 27.75 / 27.54 \\
Voxel size & \textbf{1\,cm} / 2\,cm & 23.55 / 25.11 & 27.75 / 27.75 \\
Cross-model veto & off / \textbf{on} & 19.48 / 23.55 & 25.78 / 27.75 \\
Voting & hard (argmax) / \textbf{soft} & 20.36 / 23.55 & 24.12 / 27.75 \\
\bottomrule
\end{tabular}
}
\end{table}

Table~\ref{tab:ablations} varies one component or setting at a time. Two components matter.
Replacing soft voting with a hard per-mask argmax lowers per-scene mIoU from 23.55 to 20.36 and
degrades all eight scenes (sign test $p\approx0.008$): committing per mask loses the cross-view
correction of weak 2D evidence. Turning the veto off lowers it to 19.48, because the detector label
then fires on absent classes. The remaining settings matter far less. Per-scene mIoU moves by at most
0.32 over $\alpha\in\{0.3,0.5,0.7\}$ and at most 0.62 over veto depths $k\in\{3,5,10\}$, so the method
is not sensitive to them, and 2\,cm voxels raise per-scene mIoU to 25.11 at unchanged pooled mIoU. The coarser grid also raises
AP@25 from 27.3 to 30.2 and top-1 retrieval from 24.8\% to 26.2\%, at a small cost in purity (52.4 to
51.7), and stores about 2.6 times fewer points. We keep 1\,cm for the Replica results in this paper,
which leaves them slightly below what 2\,cm gives. On
ScanNet++ the veto has the opposite trade-off: it removes absent classes but costs 1.75 present-class
mIoU (42.32 vs.\ 44.07), which is why we also report the per-dataset variant. These ablations isolate
the mechanism on Replica; the ScanNet++ controls in Sec.~\ref{sec:scannetpp} cover the larger margin.

\subsection{Analysis}\label{sec:decomp}
\paragraph{Room scale is label-limited}
On Replica, monocular VOIM is level with OVO-SLAM pooled and below it per scene
(Sec.~\ref{sec:replica}). Its score is capped by 2D labeling: the detector misses small classes,
which score zero IoU however good the reconstruction is. At room
scale the SLAM backend is interchangeable, with VGGT-SLAM~\cite{maggio2025vggtslam} matching
MASt3R-SLAM at equal frame density; we do not claim this at building scale. The remaining headroom
is in open-vocabulary detection.

\paragraph{Monocular mapping at building scale}
On the ten ScanNet++ scenes, MASt3R-SLAM on the dense phone video gives VOIM 17.87 pooled and
24.78 per-scene present-class mIoU without depth or GT poses (Table~\ref{tab:scannetpp}). That is
above HOV-SG and ConceptGraphs with both, and below the RGB-D systems. Per-scene scores range from
14.39 to 40.46, and trajectory RMSE after Sim(3) alignment is 0.03--0.92\,m. None of the baselines
has a monocular mode, since all of them need depth or GT poses, so this row shows capability rather
than a head-to-head result. It ran on an L40 GPU, the other rows on an A10G; five scenes run on both
differ by under 0.4. Estimated depth is no substitute for sensor depth: across seven GT-pose
scenes, replacing sensor depth with estimated depth lowers present-class mIoU of the per-dataset
variant from $46.40$ to $27.26$, as feed-forward depth confidence collapses on wide baselines.

\begin{figure*}[t]
\centering
\includegraphics[width=\textwidth]{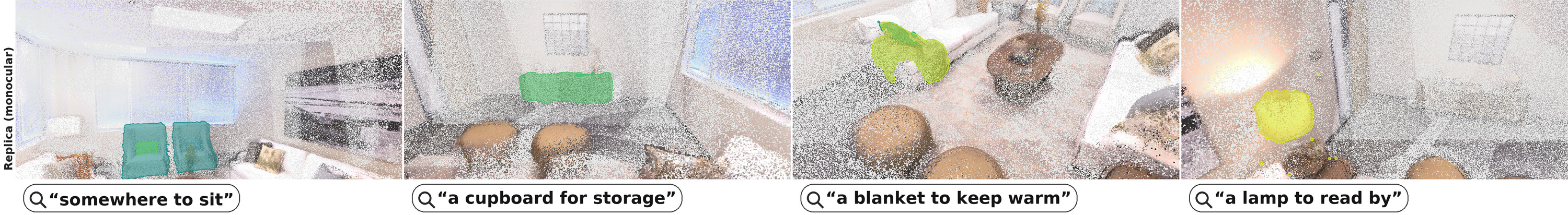}\\[3pt]
\includegraphics[width=\textwidth]{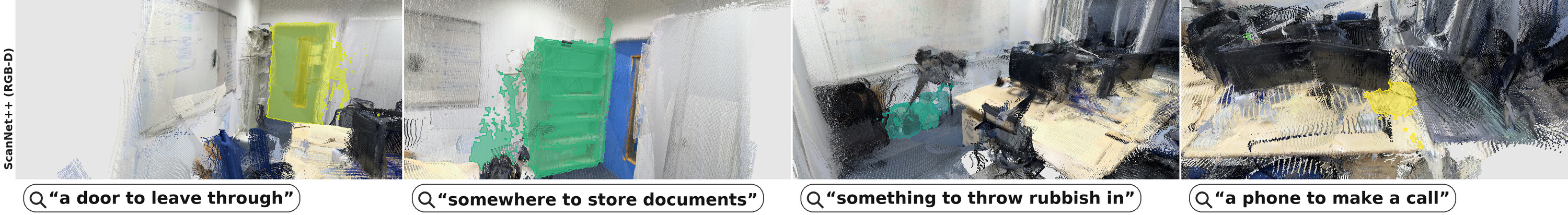}
\caption{\textbf{Open-vocabulary queries resolve to object instances.} Top: the monocular Replica
\texttt{room0} map scored in Table~\ref{tab:replica}. Bottom: ScanNet++ scene \texttt{7b6477cb95} from
RGB-D and ground-truth poses, with instance clustering enabled. Each query is matched by cosine
similarity between its text embedding and each instance's averaged vision--language embedding, not by
class name, and retrieved instances are colored by similarity, yellow most similar. Of twelve queries
written before retrieval on the ScanNet++ map, seven returned the intended object; four are shown.}
\label{fig:query}
\end{figure*}

\subsection{Runtime and memory}\label{sec:runtime}
On one NVIDIA A10G, monocular Replica room0 takes 301\,s end to end, with 13.2\,GiB peak GPU
memory. Tracking alone runs at 7.3 frames per second on room0 and 4.9 on a ScanNet++ scene.
With the semantic stream sharing the GPU, tracking slows to 4.7 and 3.7. Neither reaches the 10 and 15
frames per second assumed for these sequences, so we do not claim real-time operation. Semantic labeling runs asynchronously alongside tracking at 15.3\,s per
labeled keyframe. Over $90\%$ of that time is the per-class Grounding DINO loop over the 51-class
vocabulary, so the cost grows linearly with vocabulary size. Labeling therefore runs well below real
time, and the bottleneck is detection, not the mapping stage: the voxel accumulation adds votes
without revisiting past associations, so a single-pass open-vocabulary detector would remove most
of this cost without changing VOIM itself.

\section{Navigation-Ready Maps}\label{sec:robot}
Per-point labels project to a standard 2D occupancy grid: floor becomes free space, and an obstacle
band above the detected floor plane becomes occupied. Per-instance open-vocabulary descriptors map
free-form text to an object and a nearest goal pose. Fig.~\ref{fig:query} shows compositional
retrieval on the monocular Replica map and an RGB-D ScanNet++ map, and Fig.~\ref{fig:hero} shows the
robot at retrieved objects. On a frozen benchmark of 1,104 queries across eight Replica scenes,
top-1 retrieval reaches 24.8\% from monocular video and 32.7\% at input parity, against 23.5\% for
OVO-SLAM (MRR 0.363 and 0.450 vs.\ 0.337).

A single hand-held ZED~2i sweep of a building floor, about $30{\times}25$\,m, with sensor RGB-D, the
veto active, and the ScanNet++ top-100 vocabulary unchanged, yields 1,072 instances of 88 classes
(Fig.~\ref{fig:hero}) and a 635\,m$^2$ connected free-space region in the exported Go2 grid. Without
GT, this scene is a demonstration, not a measurement.

\subsection{Real-robot query-to-goal navigation}{\label{sec:robot_real}
\emph{Setup.} A Unitree Go2 receives a prompt fixed before any retrieval. From each of three starts
we ask for four objects, once by category name and once as an everyday request, giving 24 prompts.
Each prompt is matched to the map's per-instance index by its frozen PE-Core text embedding, taking
the top-1 instance without re-ranking. Requests are embedded as raw sentences without a prompt template,
which leaves retrieval on monocular Replica unchanged (MRR .363 vs.\ .360). A goal is then placed 0.5--1.2\,m from the instance footprint.
\emph{Object selection uses only the VOIM map.} Because the camera-SLAM map does not align exactly
with the floor's lidar map, most runs warp it onto the lidar map, using an affine plus thin-plate
spline fit with 14\,cm mean residual, and execute with Nav2 under lidar localization. The operator
judges arrival and whether the object fits the request.

\emph{Results.} Nine of twelve everyday requests ended at a fitting object (Table~\ref{tab:robot}), eight
autonomously. ``Somewhere to heat up my food'' reached a hot-water boiler (mapped as a cup); ``a way out of this room''
picked a phantom clothes hanger 2.2--2.8\,m up on a glass wall, with no reachable goal; ``where is the door'' was not
run, as the plan detoured around the glass door. These failures come from the map and the embedding, not execution: an
averaged instance embedding scores what an object is, not what it affords, and glass corrupts the map. Category names
picked the right object in all twelve runs, ten autonomously.}

\begin{table}[t]
\caption{Real-robot runs on one building floor (Go2, top-1 instance). Each object is requested
twice: as an everyday request and by its category name.}
\label{tab:robot}
\centering\scriptsize\setlength{\tabcolsep}{4pt}
\newcommand{\ok}{\checkmark}
\newcommand{\okop}{\checkmark$^{\mathrm{o}}$}
\begin{tabular}{@{}cllcc@{}}
\toprule
Start & Everyday request & Category & Request & Category \\
\midrule
\multirow{4}{*}{S0} & ``somewhere to heat up my food'' & microwave & $\sim^{a}$ & \ok \\
 & ``where can I throw this away'' & rubbish bin & \ok & \ok \\
 & ``a way out of this room'' & printer$^{b}$ & $\times^{c}$ & \ok \\
 & ``a surface to prepare food'' & kitchen counter & \ok$^{d}$ & \ok \\
\midrule
\multirow{4}{*}{S1} & ``I want to print something'' & printer & \ok & \ok \\
 & ``a table for a meeting'' & table & \okop & \okop \\
 & ``somewhere to store my files'' & cabinet & \ok$^{d}$ & \okop \\
 & ``where is the door'' & door & --$^{e}$ & \ok \\
\midrule
\multirow{4}{*}{S2} & ``a place to find cold water'' & fridge & \ok & \ok \\
 & ``somewhere to wash my hands'' & sink & \ok & \ok \\
 & ``a comfortable place to sit'' & sofa & \ok & \ok \\
 & ``a place to put my rubbish'' & rubbish bin & \ok & \ok \\
\midrule
\multicolumn{3}{@{}l}{Fitting object reached} & 9/12 & 12/12 \\
\bottomrule
\end{tabular}\\[3pt]
\parbox{\columnwidth}{\scriptsize \ok\ fitting object reached autonomously; \okop\ reached with the
operator judging arrival or steering one turn; $\sim$ useful place, wrong object; $\times$ no
reachable goal; -- not run.
$^{a}$A hot-water boiler, mapped as a cup. $^{b}$Run in place of the planned ``exit''. $^{c}$A
phantom instance high on a glass wall. $^{d}$The picked instance (a kettle; a tissue box) lies on the
fitting object. $^{e}$The plan detoured around a glass door.}
\end{table}

\section{Limitations}\label{sec:limits}
Detectors often miss ``stuff'' classes such as wall and floor; fixing this needs a dense
open-vocabulary segmenter, and the gap weakens floor-based traversability. Instance results are mixed
at input parity, with higher AP@25 and coverage but lower AP@50 and purity (Table~\ref{tab:instances}),
and monocular coverage trails the baseline for surfaces, small objects, and furniture alike. The
mapping advantage holds for ScanNet++ present-class scoring. Because VOIM predicts many classes
that are absent from a scene, OVO-SLAM leads under all-classes scoring (24.86 vs.\ 16.95); on Replica, input parity is on par with
OVO-SLAM per scene and monocular VOIM trails it. Monocular maps stay below RGB-D maps
(Sec.~\ref{sec:decomp}), estimated depth degrades on wide baselines, and monocular scale must come
from an external source such as calibrated stereo or an IMU; without one, no monocular result here
is metric. Per-class detection scales linearly with vocabulary size, the semantic stream lags behind
tracking and runs well below real time (Sec.~\ref{sec:runtime}), and a linear per-point feature buffer runs out at 23\,GiB near 4\,M 1024-D
points. Finally, a language-queryable index of a home is also a searchable record of a private space.

Code, configurations, the evaluation scripts (Sec.~\ref{sec:protocol}), frame lists, scene IDs,
and per-scene score CSVs will be released upon acceptance.

\section{Conclusion}\label{sec:conclusion}
VOIM is a training-free instance mapper that accumulates soft evidence in every voxel across views
before deciding labels and instances. On ScanNet++ with RGB-D and ground-truth poses, it improves
present-class mIoU over OVO-SLAM, HOV-SG, and ConceptGraphs in all ten scenes, by 9.95 per scene and
8.01 pooled over the strongest of them, and its detector-free variant still wins nine of ten. On
Replica it is ahead pooled and on par per scene. From monocular video alone, it exceeds HOV-SG and
ConceptGraphs run with depth and GT poses on ScanNet++. Its per-instance index improves query retrieval,
and on a real building floor the map led a quadruped to a fitting object for nine of twelve everyday
requests.

\bibliographystyle{IEEEtran}
\begingroup\let\footnotesize\scriptsize
\bibliography{references}
\endgroup
\end{document}